\documentclass[10pt,twocolumn,letterpaper]{article}
\usepackage{arxiv}

\usepackage[dvipsnames]{xcolor}

\usepackage[font={footnotesize}]{caption}
\usepackage{diagbox}
\usepackage{multirow}
\usepackage{wrapfig}
\usepackage{graphicx}
\usepackage{amsmath}
\usepackage{amssymb}
\usepackage{booktabs}
\usepackage{cite}

\usepackage[flushleft]{threeparttable}

\usepackage{times}
\usepackage{epsfig}
\usepackage{graphicx}
\usepackage{amsmath}
\usepackage{amssymb}

\definecolor{cvprblue}{rgb}{0.21,0.49,0.74}
\usepackage[pagebackref,breaklinks,colorlinks,citecolor=cvprblue]{hyperref}

\title{Hide and Seek: How Does Watermarking Impact Face Recognition?}

\author{Yuguang Yao$^*$\quad\quad Steven Grosz$^*$\quad\quad Sijia Liu\quad\quad Anil Jain\\
Michigan State University}

\begin{document}

\maketitle

\def\thefootnote{*}\footnotetext{Equal contribution}\def\thefootnote{\arabic{footnote}}

\begin{abstract}
The recent progress in generative models has revolutionized the synthesis of highly realistic images, including face images. This technological development has undoubtedly helped face recognition, such as training data augmentation for higher recognition accuracy and data privacy. However, it has also introduced novel challenges concerning the responsible use and proper attribution of computer generated images. We investigate the impact of digital watermarking, a technique for embedding ownership signatures into images, on the effectiveness of face recognition models. We propose a comprehensive pipeline that integrates face image generation, watermarking, and face recognition to systematically examine this question. The proposed watermarking scheme, based on an encoder-decoder architecture, successfully embeds and recovers signatures from both real and synthetic face images while preserving their visual fidelity.  Through extensive experiments, we unveil that while watermarking enables robust image attribution, it results in a slight decline in face recognition accuracy, particularly evident for face images with challenging poses and expressions. Additionally, we find that directly training face recognition models on watermarked images offers only a limited alleviation of this performance decline. Our findings underscore the intricate trade off between watermarking and face recognition accuracy. This work represents a pivotal step towards the responsible utilization of generative models in face recognition and serves to initiate discussions regarding the broader implications of watermarking in biometrics.
\end{abstract}

\newpage
~
\vspace{60mm}
\section{Introduction}
\label{sec: intro}

Recent advancements in generative modeling, particularly in diffusion models (DMs) \cite{saharia2022photorealistic,rombach2022high,ramesh2022hierarchical}, have demonstrated remarkable capabilities in synthesizing complex and highly realistic images. However, as this technology evolves, it brings forth new considerations regarding its safety and ethical use. These emerging issues include harmful content generation \cite{schramowski2023safe,zhang2023generate},  deep fakes \cite{agarwal2019protecting,chesney2019deep}, privacy leakage \cite{li2024shake,carlini2023extracting},  copyright usurpation \cite{somepalli2023understanding,somepalli2023diffusion,zhang2024unlearncanvas},  and societal biases and stereotyping \cite{luccioni2024stable,jiang2023rs}. Therefore, significant efforts have been made to avoid the misuse of generative models and to enhance their safety, security, and trustworthiness \cite{fan2023trustworthiness}.  For example, the \textit{watermarking} technology \cite{jain2003hiding,zhu2018hidden,fernandez2023stable,wen2023tree}, which we will focus on in this work,  entails embedding a distinctive signature into image generation.  This approach enables differentiation between real and synthesized images and can be used to protect image copyrights and ascertain ownership.

\begin{figure}[t]
\centering
\includegraphics[width=\columnwidth,height=!]{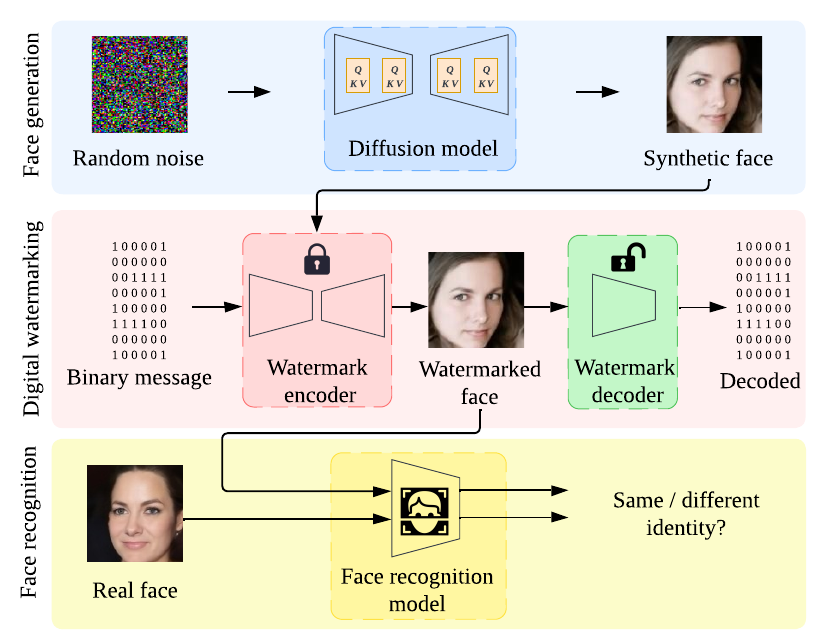}
\vspace*{-2mm}
\caption{{
The schematic overview of the studied face recognition pipeline, which
 consists of face image generation (top), digital watermarking (middle), and face recognition (bottom). The watermarking technique follows HiDDeN \cite{zhu2018hidden}.
}}
 \vspace*{-3mm}
\label{fig: overview}
\end{figure}

Among the myriad applications of generative models, \textit{face image generation} has garnered considerable attention recently. For example, the generation of high-fidelity synthetic face images enables model developers to train face recognition systems solely using synthetic data, thereby circumventing the risk of compromising the privacy of real individuals \cite{kim2023dcface}. 
Augmenting existing training sets with synthesized face images can also help mitigate racial biases and tackle data imbalance problems \cite{yi2014learning,zhu2021webface260m}.
Furthermore, generative models have been employed to manipulate or enhance face images \cite{shi2019warpgan,ye2023ip,li2023photomaker,wang2024instantid}. Recent advances in text to image generation using denoising diffusion probabilistic models (DDPM) have allowed for synthesizing highly realistic, novel views and appearances of a given face image, while preserving the identity. 

Despite the extensive use of generative models in face recognition, there has been limited research delving into the implications of integrating watermarking into these models and how their watermarked image generations influence their efficacy in face recognition tasks.
In particular, we ask:

\begin{center}
    \textit{\textbf{(Q)} 
How effective is face recognition when evaluated on watermarked face images?
}
\end{center}

\noindent To address \textbf{(Q)}, we explore the influence of existing digital watermarking technology in face recognition.
The proposed system, as shown in  \textbf{Fig.\,\ref{fig: overview}}, comprises a diffusion model (DM) for face image generation, an encoder-decoder-based network for embedding a watermark signature into synthetic images, and a face recognition model trained on watermarked data. Assessing through face verification tasks, aimed at determining whether two face images belong to the same person, we observe that watermarking retains effectiveness throughout the generation and recognition processes. Nonetheless, we do notice a minor decline in face recognition accuracy, particularly noticeable when comparing challenging face images.

\paragraph{Contributions.} We summarize our contributions below.

\noindent 
$\bullet$ We make a novel exploration into the impact of watermarked face images on face recognition accuracy.

\noindent 
$\bullet$ We investigate face recognition through a unified framework that integrates watermarking in face image generation. This facilitates us to train and evaluate face recognition systems with watermarked synthetic data.

\noindent 
$\bullet$ Through extensive experiments, we showcase the effectiveness of watermarking in face generation and recognition. We highlight its resilience against post-hoc image transformations and offer valuable insights into the successes and challenges encountered when employing watermarked images for face recognition training.

\section{Related Work}

\paragraph{Face image generation.} 
Recent advancements in generative modeling have revolutionized the synthesis of high-quality face images. Generative Adversarial Networks (GANs) \cite{goodfellow2014generative} have been extensively employed for face image generation, exemplified by notable architectures like StyleGAN \cite{karras2019style} and StarGAN \cite{choi2018stargan}. These models adeptly translate random noise vectors into realistic face images, facilitating the creation of diverse and visually appealing facial representations. However, GANs often encounter issues with training instability and mode collapse, hindering their capacity to capture the full spectrum of facial variations.

More recently, diffusion models \cite{song2019generative,ho2020denoising} have emerged as a compelling alternative for high-fidelity image generation. Operating by iteratively denoising a Gaussian noise signal and guided by a learned noise schedule, diffusion models excel in generating highly realistic and varied face images, surpassing GANs in many respects. Prominent examples of diffusion-based face generation models include DDPM \cite{ho2020denoising} and DCFace \cite{kim2023dcface}.
The advent of sophisticated face image generation techniques has introduced new avenues for data augmentation, privacy preservation, and creative applications \cite{cao2018vggface2,karras2020training,mirjalili2018semi}. However, it has also raised concerns regarding potential misuse, such as the creation of deepfakes \cite{westerlund2019emergence} or identity theft. These apprehensions underscore the necessity for methods ensuring the responsible use and attribution of computer generated face images.

\paragraph{Digital watermarking.}
Digital watermarking is a technique for embedding hidden information, known as a watermark, into digital content. While the watermark can be designed to be visible or invisible, we deal with watermarks which are imperceptible to human observers but can be detected and extracted by specialized algorithms. Digital watermarking has a wide range of applications, including copyright protection, content authentication, and data provenance \cite{cox2007digital,miller2000informed,voloshynovskiy2001attacks,fridrich2000visual}. Early works on digital watermarking focused on developing techniques that relied on transform-domain techniques, such as discrete cosine transform (DCT) \cite{piva1997dct,barni1998dct}, discrete wavelet transform (DWT) \cite{xia1997multiresolution}, or singular value decomposition (SVD) \cite{liu2002svd}, to embed the watermark in the frequency or singular value domain of the image.

With the advent of deep learning, researchers have explored the use of neural networks for digital watermarking. These learning-based approaches can automatically learn to embed and extract watermarks from images, potentially offering improved robustness and capacity compared to traditional methods. Notable examples include HiDDeN \cite{zhu2018hidden}, which uses an encoder-decoder architecture to hide watermarks in images, and DeepStego \cite{zhang2019steganogan}, which employs adversarial learning to enhance the imperceptibility of the watermark. 
There have also been works studying watermarking for GAN images  \cite{yu2021artificial,zhang2021deep,skripniuk2020watermarking,wu2020watermarking,fei2022supervised}.
However, the impact of watermarking on face recognition has not been well studied. Our work aims to address this gap by investigating the interplay between digital watermarking and face recognition.

\paragraph{Face recognition.}
Face recognition technology has garnered substantial attention in computer vision research due to its diverse applications in security, surveillance, and human-computer interaction. 
Early face recognition methods predominantly relied on handcrafted features and classic machine learning algorithms. Techniques such as Eigenfaces~\cite{turk1991eigenfaces}, Local Binary Patterns~\cite{rahim2013face}, and Histogram of Oriented Gradients~\cite{albiol2008face} have been widely used for feature extraction, whereas classifiers like Support Vector Machines~\cite{dadi2016improved} and k-Nearest Neighbors~\cite{parveen2006face} were used for reocgnition.

The advent of deep learning revolutionized face recognition, where Convolutional Neural Networks (CNNs), such as DeepFace by Taigman et al.~\cite{taigman2014deepface}, enabled significant improvements in accuracy and robustness. Since then, many different types of loss functions (\textit{e.g.}, SphereFace~\cite{liu2017sphereface}, CosFace~\cite{liu2017sphereface}, ArcFace~\cite{deng2019arcface}, etc.) and model architectures (e.g., VGG~\cite{simonyan2014very}, ResNet~\cite{he2016deep}, ViT~\cite{dosovitskiy2020image}, etc.) for large-scale face recognition were proposed, further pushing the boundaries of face recognition. These algorithms relied on the availability of large-scale face recognition datasets for their success. Yet, many of these datasets, such as the popular MS-Celeb-1M~\cite{guo2016ms}, have now been recalled due to privacy concerns. This has prompted many researchers to turn toward developing methods for realistic face image generation to replace these large-scale face recognition datasets.

\section{Methodology}

To tackle the question \textbf{(Q)} posed in Sec.\,\ref{sec: intro}, this section introduces the watermarking, face generation, and face recognition techniques utilized in constructing the proposed system, as depicted in Fig.\,\ref{fig: overview}.

\subsection{Digital Watermarking}
\label{sec: watermarking_method}
Digital watermarking embeds hidden information, known as a `watermark' or `signature', into digital content like images. Such watermarks are often imperceptible to humans but detectable by specialized algorithms. In the case of (synthesized) face images, watermarking serves to track their origin and deter unauthorized usage or distribution \cite{jain2003hiding,zhu2018hidden,fernandez2023stable,wen2023tree}.

To be concrete, the watermarking problem can be defined as follows: Given an input image $\mathbf{I} \in \mathbb{R}^{H \times W \times C}$ and an $L$-bit watermark message $\mathbf{m} \in \{0, 1\}^L$ (\textit{i.e.}, digital signature), our goal is to produce a watermarked image $\mathbf{I}_{\mathrm{w}}$ that maintains visual similarity to the original image $\mathbf{I}$ while incorporating the watermark message $\mathbf{m}$. In addition, the watermarked image $\mathbf{I}_{\mathrm{w}}$ should be designed such that it allows for the reverse engineering of its encoded message $\mathbf m$, enabling the deciphering of the image's origin. 

\begin{figure}[htb]
\centering
\includegraphics[width=0.8\columnwidth,height=!]{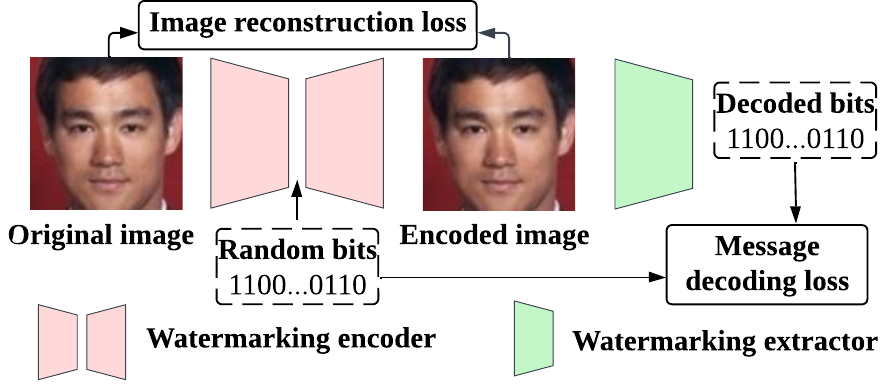}
\caption{{
The schematic overview of watermarking system.  
}}
\label{fig: watermarking}
\end{figure}

To implement the watermarking system, we employ the neural network-based 
HiDDeN framework \cite{zhu2018hidden} due to its proven performance while any other watermarking scheme can be used as well; see Fig.\,\ref{fig: watermarking} for a schematic overview. This watermarking system consists of an encoder network $f_{\boldsymbol \theta}$ and a decoder network $g_{\boldsymbol \phi}$. The encoder takes the input image $\mathbf{I}$ and the watermark message $\mathbf{m}$ as inputs and generates the watermarked image $\mathbf{I}_{\mathrm{w}} = f_{\boldsymbol \theta}(\mathbf{I}, \mathbf{m})$. The decoder takes the watermarked image $\mathbf{I}_{\mathrm{w}}$ as input and reconstructs the embedded watermark message $\hat{\mathbf{m}} = g_{\boldsymbol \phi}(\mathbf{I}_{\mathrm{w}})$.

The encoder and decoder networks are jointly trained using a combination of image reconstruction loss and message decoding loss.  
The image reconstruction loss $\ell_{\mathrm{recons}}$  (\textit{e.g.}, mean squared error)
ensures that the watermarked image is visually similar to the original image, while the message decoding loss
$\ell_{\mathrm{decode}}$ (\textit{e.g.}, bit-wise binary cross-entropy loss)
minimizes the difference between the embedded and extracted watermark messages. 
The overall training objective of the watermarking system is then given by
\begin{align}
\displaystyle \min_{\boldsymbol \theta, \boldsymbol \phi} \,\,  \mathbb E_{\mathbf I, \mathbf m} \left [ \ell_{\mathrm{recons}}(\mathbf{I}_{\mathrm{w}}, \mathbf{I}) + \lambda \ell_{\mathrm{decode}}(g_{\boldsymbol \phi}(\mathbf{I}_{\mathrm{w}}), \mathbf{m}) \right ],
\end{align}%
where recall that
$\mathbf{I}_{\mathrm{w}} = f_{\boldsymbol \theta}(\mathbf{I}, \mathbf{m})$, 
  and $\lambda$ is a regularization parameter to strike a balance between the image reconstruction and the message decoding losses.
The training dataset is given by MS-COCO \cite{lin2014microsoft}.

It is also worth noting that during training, a random message generator is employed to produce random bits in $\mathbf m$ (as depicted in Fig.\,\ref{fig: watermarking}), which are then combined with the image input at the input of the encoder network. The resilience to random bit messages allows us to utilize a pre-trained encoder-decoder-based watermarking system directly to embed a user-defined watermark message in face images, which we will employ subsequently.
In addition, the watermark training integrates data augmentations. That is, watermarked images undergo diverse transformations, including random cropping, resizing, and compression, mimicking real-world scenarios. Through training with augmented data, the encoder and decoder networks learn to embed and extract watermarks resilient to these transformations. As will be evident later, this renders it an effective approach for watermarking generated face images, ensuring the robustness of decoding the watermark against external image transformations.

\subsection{Integration of Watermark into Face Images}
\begin{figure}[htb]
\centering
\includegraphics[width=0.65\linewidth,height=!]{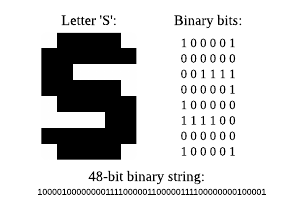}
\vspace*{-3mm}
\caption{
The embedded watermarking signature.
}
\label{fig: letter_S}
\vspace*{-3mm}
\end{figure}

We begin by incorporating a designated watermark into face images. This integration is accomplished through the encoder network ($f_{\boldsymbol \theta}$) of the trained watermarking system. As shown in Fig.\,\ref{fig: letter_S},
we select the `$S$' letter (represented as a binary image) as the watermark message, with its pixel values utilized for the binary message, denoted as $\mathbf{m}_{S}$.
Therefore, given a face image $\mathbf I$, its watermarked version is now given by 
$\mathbf{I}_{\mathrm{w}} = f_{\boldsymbol \theta} (\mathbf I, \mathbf{m}_{S})$. 

In addition to the specificity of the watermark message, we also consider two sources of face images, including both \textit{synthetic} and \textit{real} face images.
The synthetic image source is \textit{DCFace} \cite{kim2023dcface}, a state-of-the-art diffusion-based face generation model. DCFace is built upon DMs (diffusion models) to enhance the quality and controllability of generated face images conditioned on individual identities and face variations.   It excels at producing face images of a specific identity across various styles while allowing for precise control, thereby enriching face recognition models with augmented training data of large intra-class variance.  
We utilize DCFace to simulate the scenario where synthesized face images require integration with the watermark $\mathbf{m}_{S}$ before their downstream applications, like face recognition.
The second image source consists of commonly-used real face image datasets, such as \textit{CASIA-WebFace}~\cite{yi2014learning} and \textit{MS-Celeb-1M v2 (MS1Mv2)}~\cite{deng2019arcface}, which include face images of real identities. We will apply watermarking to both image sources respectively and investigate its effects on face recognition.

\begin{figure}[htb]
\includegraphics[width=\linewidth,height=!]{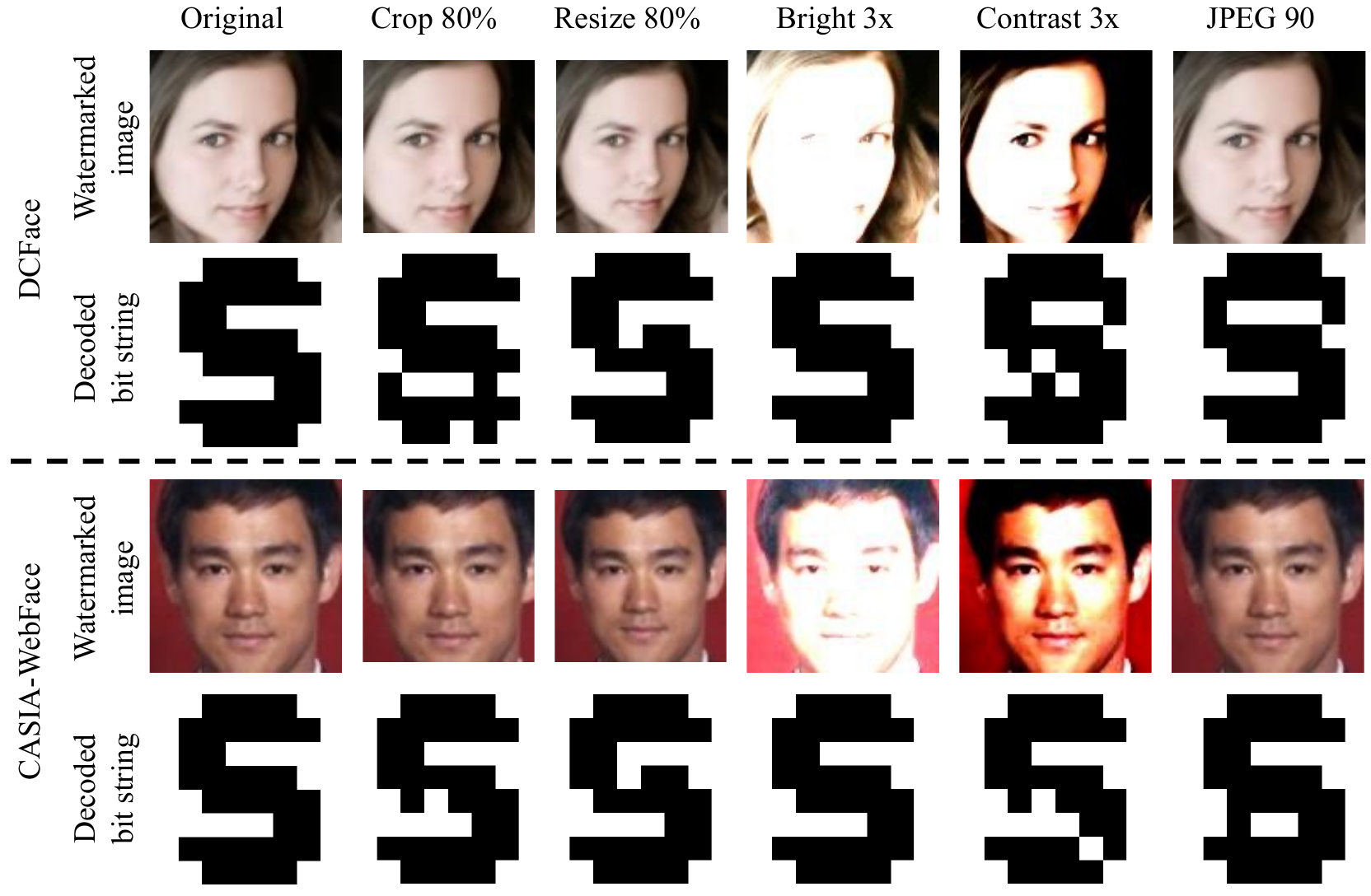}
\vspace*{-2mm}
\caption{{
Visualizations of watermarked images and their decoded watermark message (the `$S$' letter) based on the real image source CASIA-WebFace and the synthetic image source DCFace. Each column represents a data transformation operation applied after image watermarking. `Original' denotes no data transformation. `Crop 80\%' refers to randomly cropping the image to 80\% of its original size. `Resize 80\%' refers to resizing the image to 80\% of its original size. 
`Bright 3x' refers to increasing the image brightness by a factor of 3. 
`Contrast 3x' refers to increasing the image contrast by a factor of 3. 
`JPEG 90\%' refers to applying JPEG compression with a quality factor of 90\%.
}}
\vspace*{-3mm}
\label{fig: transformation}
\end{figure}

The performance of image watermarking is evaluated by assessing the reconstruction accuracy of the watermark message using the decoder network $g_{\boldsymbol \phi}$ applied to the watermarked image $\mathbf{I}_{\mathrm{w}}$.
In particular, the watermark is expected to exhibit robustness against post-hoc image transformations, 
ensuring the reliable extraction of the watermark message.
\textbf{Fig.\,\ref{fig: transformation}} shows the effectiveness of image watermarking against a sequence of post-watermarking data transformations under both  the synthetic image source DCFace and the real image source CASIA-WebFace.


\subsection{Face Recognition on Watermarked Images} 
With a dataset of watermarked face images in hand, we proceed to conduct the face recognition task. 
In this study, our face recognition setup adheres to AdaFace \cite{kim2022adaface}, while any other face recognition algorithm can be used as well. 
 Given an input face image $\mathbf{I}$,  a face recognition model $h_{\boldsymbol \psi}$ maps the image to a compact feature representation $\mathbf{z}$: 
 $
 \mathbf{z} = h_{\boldsymbol  \psi} (\mathbf I)
 $,
where $\boldsymbol{\psi}$ represents the learnable parameters of the model. The feature representation $\mathbf{z}$ is typically obtained from the penultimate layer of a CNN, such as ResNet \cite{he2016deep} or MobileFaceNet \cite{chen2018mobilefacenets}.
The model is trained to minimize a classification loss, such as the softmax loss \cite{deng2019arcface} or the margin-based loss \cite{liu2017sphereface,wang2018cosface,deng2019arcface}, which reduces the intra-class variance in the embedding space while increasing the inter-class variance.

During inference, the learned face recognition model extracts the feature representation $\mathbf{z}_{\mathrm{p}}$ for a probe face image $\mathbf{I}_{\mathrm p}$. The similarity between the probe face $\mathbf{I}_{\mathrm p}$ and a reference face $\mathbf{I}_{\mathrm{r}}$ (with feature representation $\mathbf{z}_\mathrm{r}$) is computed using a similarity metric, such as the cosine similarity
\begin{align}
s(\mathbf{z}_{\mathrm{p}}, \mathbf{z}_{\mathrm{r}}) = \frac{\mathbf{z}_{\mathrm{p}}^T \mathbf{z}_{\mathrm{r}}}{|\mathbf{z}_{\mathrm{p}}| |\mathbf{z}_{\mathrm{r}}|},
\end{align}
where $|\cdot|$ denotes the Euclidean norm. The probe face is considered to match the reference face if their similarity score exceeds a predefined threshold $\tau$:
\begin{align}
\text{match}(\mathbf{z}_{\mathrm{p}}, \mathbf{z}_{\mathrm{r}}) = 
\begin{cases}
1, & \text{if } s(\mathbf{z}_{\mathrm{p}}, \mathbf{z}_{\mathrm{r}}) \geq \tau, \\
0, & \text{otherwise}.
\end{cases}
\end{align}

The learned face recognition models are commonly assessed through their verification (1:1 comparison) performance. For verification, metrics such as true acceptance rate (TAR) at a fixed false acceptance rate (FAR) are reported. In this paper, we report the TAR at $\mathrm{FAR}=0.01\%$.

Given the above system and protocol, the next section will empirically demonstrate the impact of watermarking on the performance of face recognition models trained using watermarked face images.





\begin{figure*}[htb]
\includegraphics[width=\textwidth,height=!]{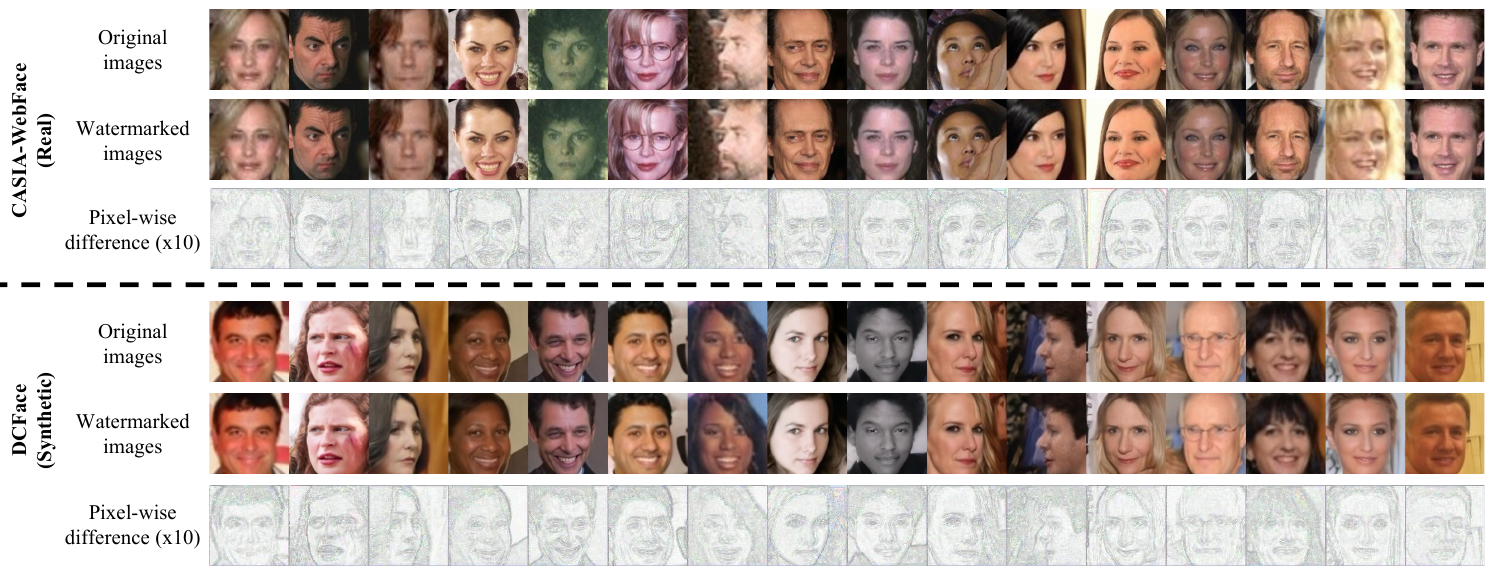}
\caption{\footnotesize{
{Visualization of original images, watermarked images, and their 10$\times$ pixel differences for both real and synthetic face image datasets, CASIA-WebFace \cite{yi2014learning} and DCFace \cite{kim2023dcface}, respectively.
The pixel-wise difference is color-inverted for enhanced visualization, where white areas denote zero values.
}
}}
\label{fig: visualization_watermarking}
\end{figure*}

\section{Experiments}
In this section, we demonstrate the quality of watermarked face images and their robustness to diverse image transformations. Next, we assess the impact of watermarking face images on the performance of  given, pre-trained face recognition models. Finally, we investigate the effect of using watermarked face images to train face recognition models on downstream face recognition tasks.



\begin{table}[htb]
\caption{{Summary of face recognition datasets used in this study.}}
\centering
\resizebox{0.7\linewidth}{!}{%
\begin{tabular}{lccc}
\toprule
\midrule
\textbf{Dataset} & \textbf{\# of IDs} & \textbf{\# of Images} & 
\begin{tabular}{c}
    \textbf{Avg. \#}  \\
     \textbf{(Images/ID)} 
\end{tabular}
\\
\midrule
MS1Mv2~\cite{deng2019arcface} & 85,742 & 5.8M & 68 \\
\midrule
CASIA-WebFace~\cite{yi2014learning} & 10,572     & 0.5M & 46 \\
\midrule
DCFace~\cite{kim2023dcface} & 60,000 & 1.3M & 22 \\
\midrule
\bottomrule
\end{tabular}
}
\label{tab: datasets}
\end{table}

\subsection{Experimental Setup}

First, in \textbf{Table\,\ref{tab: datasets}}, we provide a description of the datasets utilized in this study, including the number of unique face identities, the total number of images, and the average number of images per identity. For evaluation purposes, we select a subset of 2 images per identity from all 10,547 identities in CASIA-WebFace and 2 images per identity from a randomly sampled set of 10,000 identities for DCFace to compute genuine and imposter scores. During training experiments, we utilize the full MS1Mv2 dataset.

Second, we implement the watermarking system following Sec.\,\ref{sec: watermarking_method} and
the architecture setup of HiDDeN \cite{zhu2018hidden}. The watermark encoder consists of 4 Conv-BN-ReLU blocks, each with 64 conv filters of 3$\times$3 kernels, 1 stride, and 1 padding. The decoder has 7 such blocks, followed by another block with $L$ filters ($L$ is the watermarking bit string length) and a linear layer of $L$$\times$$L$. The watermark encoder and decoder are pre-trained together on the whole MS-COCO dataset \cite{lin2014microsoft}, cropped to $256 \times 256$. The watermarking performance will be evaluated on both CASIA-WebFace~\cite{yi2014learning} and DCFace~\cite{kim2023dcface} datasets.

Furthermore, to assess the impact of watermarking on face recognition model performance, we employ three state-of-the-art face recognition models pretrained on large, {non-watermarked} face recognition datasets. Specifically, we consider two variations (ResNet50 and ResNet101) of AdaFace \cite{kim2022adaface} pretrained on MS1Mv2, along with one commercial-off-the-shelf (COTS) face recognition SDK\footnote{This COTS SDK is among the top performers in the NIST FRVT Ongoing competition. \url{https://pages.nist.gov/frvt/html/frvt11.html}}.
We evaluate the performance on both the original CASIA-WebFace and DCFace datasets, as well as their watermarked versions.
For evaluation, we denote the face recognition performance on the original dataset as `\textit{original-original}', indicating that both the probe and reference face images being compared are original images. `\textit{Watermarked-original}' denotes that the probe image is watermarked while the reference image is unmodified, and `\textit{watermarked-watermarked}' indicates that both the probe and reference images are watermarked.



Finally, to explore the impact of watermarking on face recognition model training, we train our own ResNet50 AdaFace models on both the original MS1Mv2 dataset and its watermarked version. We then compare the performance of these trained models on the CASIA-WebFace dataset. Similarly, we conduct the same experiment by training on both the original DCFace synthetic face image dataset and its watermarked version to investigate the effect of watermarking on synthetic face images.

\subsection{Experimental Results}

\paragraph{Assessing watermarking quality in face images.} 
In \textbf{Fig.\,\ref{fig: visualization_watermarking}}, we present visualizations of original images, watermarked images, and pixel-wise differences between the original and watermarked images for both real and synthetic face image datasets (CASIA-WebFace and DCFace, respectively). Notably, the watermarked images exhibit no visually perceptible differences compared to the originals, with pixel-wise discrepancies primarily located along image edges, which are challenging for humans to discern.

\begin{figure}[htb]
\vspace*{-3mm}
\centering
\includegraphics[width=0.55\linewidth,height=!]{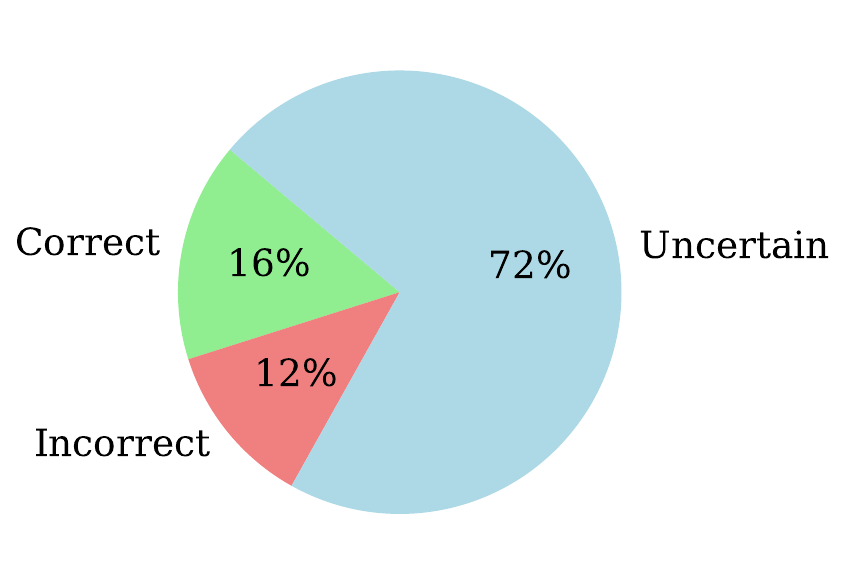}
\vspace*{-5mm}
\caption{
Survey for differentiating original/watermarked images.
}
\label{fig: survey}
\end{figure}

To further evaluate this, we randomly selected 5 pairs of original/watermarked images and solicited feedback from 10 student volunteers to differentiate between them.  
Among the total 50 responses received, 8 were correct selections, 6 were incorrect selections, and 36 were labeled as `Not sure' (See \textbf{Fig.\,\ref{fig: survey}}). This underscores the invisibility of the watermark (the `$S$' letter used in Fig.\,\ref{fig: letter_S}) to human eyes, indicating the stealthiness of watermarking and its high quality.

\textbf{Table\,\ref{tab: robustness_transformation}} extends the analysis from Fig.\,\ref{fig: transformation}, evaluating the watermark's resilience against different post-watermarking transformations (cropping, resizing, brightness, contrast, and JPEG compression) at varying scaling factors. For each transformation, scaling factors range from 1 to 0.75 for cropping and resizing, 1 to 3.5 for brightness and contrast adjustments, and 100 to 75 for JPEG compression quality. As we can see, reconstructing the watermark maintains high bit accuracy across diverse transformations and scaling factors for both real (CASIA-WebFace) and synthetic (DCFace) datasets. Notably, even when images are scaled down to 75\% of their original size, bit accuracy remains above 84\% for cropping and resizing. Likewise, for brightness and contrast adjustments, bit accuracy remains above 88\% up to a scaling factor of 3.5. However, JPEG compression has a more pronounced impact on watermark extraction, with bit accuracy dropping to approximately 73\% and 68\% for CASIA-WebFace and DCFace, respectively, at a quality factor of 75.

\begin{table}[htb]
\centering
\caption{{
{
The robustness of watermarking evaluated using the reconstructed watermark bit accuracy (\%) against various data transformations at different scaling strengths. Each value is averaged over 2000 images from each dataset, with an image size of 112 $\times$ 112 and a watermark string bit length of 48.
}
}}
\resizebox{0.95\linewidth}{!}{%
\begin{tabular}{cccccccc}
\toprule[1pt]
\midrule
\multirow{2}{*}{\textbf{Transformation}} & \multirow{2}{*}{\textbf{Dataset}} & \multicolumn{6}{c}{\textbf{Scaling ratio}}\\
\cmidrule(lr){3-8}
& & 1 & 0.95 & 0.9 & 0.85 & 0.8 & 0.75\\
\midrule
\multirow{2}{*}{Crop} & CASIA-WebFace & 98.39 & 97.22 & 93.7 & 95.12 & 94.77 & 94.3\\
& DCFace & 97.31 & 96.27 & 92.37 & 93.96 & 93.48 & 93.25\\
\midrule
\multirow{2}{*}{Resize} & CASIA-WebFace & 98.39 & 92.47 & 92.0 & 91.58 & 89.62 & 85.93\\
& DCFace & 97.31 & 90.24 & 89.99 & 89.64 & 87.09 & 84.32\\
\midrule
\midrule
\multirow{2}{*}{\textbf{Transformation}} & \multirow{2}{*}{\textbf{Dataset}} & \multicolumn{6}{c}{\textbf{Scaling factor}}\\
\cmidrule(lr){3-8}
& & 1 & 1.5 & 2 & 2.5 & 3 & 3.5\\
\midrule
\multirow{2}{*}{Brightness} & CASIA-WebFace & 98.39 & 98.48 & 96.65 & 94.21 & 91.6 & 88.87\\
& DCFace & 97.31 & 98.55 & 96.66 & 94.15 & 91.3 & 88.38\\
\midrule
\multirow{2}{*}{Contrast} & CASIA-WebFace & 98.39 & 98.81 & 98.15 & 96.82 & 94.92 & 92.62\\
& DCFace & 97.31 & 98.87 & 98.76 & 98.14 & 97.11 & 95.62\\
\midrule
\midrule
\multirow{2}{*}{\textbf{Transformation}} & \multirow{2}{*}{\textbf{Dataset}} & \multicolumn{6}{c}{\textbf{JPEG quality factor}}\\
\cmidrule(lr){3-8}
& & 100 & 95 & 90 & 85 & 80 & 75\\
\midrule
\multirow{2}{*}{JPEG compression} & CASIA-WebFace & 98.39 & 90.36 & 85.0 & 80.8 & 76.65 & 73.06\\
& DCFace & 97.31 & 87.65 & 80.02 & 75.34 & 71.75 & 68.53\\
\midrule
\bottomrule[1pt]
\end{tabular}
}
\label{tab: robustness_transformation}
\end{table}

\begin{figure*}[htb]
\includegraphics[width=\linewidth,height=!]{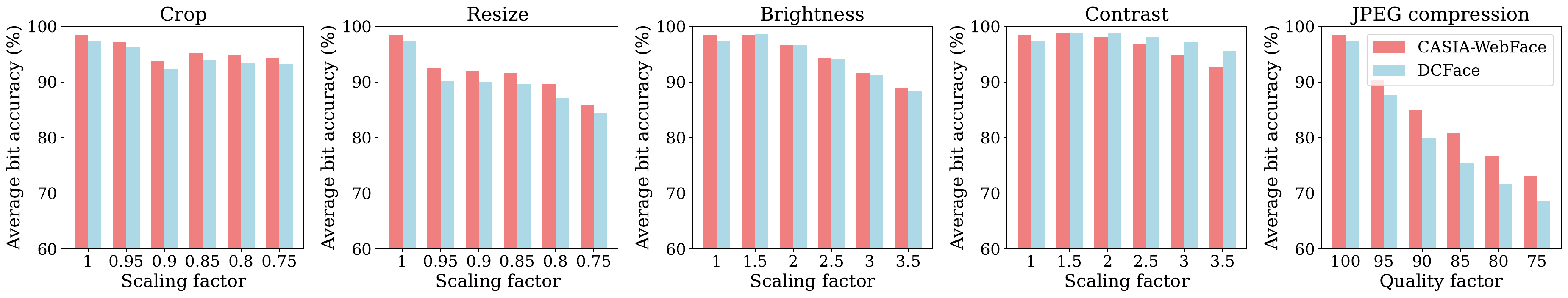}
\caption{\footnotesize{
{Watermarking robustness to different data transformations across various transformation factors. The setup follows Table\,\ref{tab: robustness_transformation}.}
}}
\label{fig: transformation_ratio}
\end{figure*}

\begin{table*}[htb]
\caption{{Pre-trained face recognition performance, TAR (\%) at FAR=0.01\%, on original images vs. watermarked images.
In the prob-preference pairs, `\textit{original-original}' indicates that both probe and reference face images to be compared are original images. `\textit{Watermarked-original}' represents that the probe image is watermarked while the reference image is unmodified, and `\textit{watermarked-watermarked}' signifies that both the probe and reference images are watermarked.
}}
\begin{center}
\resizebox{0.8\linewidth}{!}{%
\begin{threeparttable}
\begin{tabular}{l|c|ccc}
\toprule[1pt]
\midrule
\textbf{Evaluation Dataset} & \textbf{Probe-Reference} & \textbf{AdaFace ResNet50}\tnote{1} & \textbf{AdaFace ResNet101}\tnote{1} & \textbf{COTS SDK}
\\
\midrule
\multirow{3}{*}{\begin{tabular}[c]{@{}l@{}}CASIA-WebFace\\ (10,572 IDs, 2 images/ID)\end{tabular}} & Original-Original & \textbf{86.94} & \textbf{87.85} & \textbf{89.22} \\
& Watermarked-Watermarked & 86.36 & 87.62 & 88.68     \\
& Watermarked-Original & 86.61 & 86.61 & 88.66     \\
\midrule
\multirow{3}{*}{\begin{tabular}[c]{@{}l@{}}DCFace\\ (10,000 IDs, 2 images/ID)\end{tabular}} & Original-Original & \textbf{76.10} & \textbf{74.99} & \textbf{86.84}     \\
& Watermarked-Watermarked & 73.65 & 71.83 & 84.56     \\
& Watermarked-Original & 74.83 & 74.83 & 85.47    \\
\midrule
\bottomrule[1pt]
\end{tabular}
\begin{tablenotes}
\item \tnote{1}pre-trained on MS1MV2 dataset
\end{tablenotes}
\end{threeparttable}
}
\end{center}
\vspace{-5mm}
\label{tab: pretrained_results}
\end{table*}


\textbf{Fig.\,\ref{fig: transformation_ratio}} depicts the averaged bit accuracy of watermark reconstruction from watermarked face images, based on the CASIA-WebFace and DCFace datasets respectively, across different scaling factors for each data transformation type. Overall, the bit accuracy on DCFace is slightly lower than that on CASIA-WebFace, likely due to differences in image characteristics between real and synthetic faces, given that the watermarking model is pre-trained on real images. Nevertheless, the watermarking scheme demonstrates robust resilience to various transformations on both datasets. The consistently high bit accuracy across different scaling factors indicates that the watermark can be reliably extracted even when watermarked images undergo common image processing operations. This underscores the robustness and practicality of the watermarking approach for protecting and authenticating face images.

\paragraph{Watermarking effect on pre-trained face recognition models.}
Utilizing three SOTA face recognition systems, including two open-source variants (ResNet50 and ResNet101 variants of AdaFace \cite{kim2022adaface}), and a third from a COTS face recognition SDK, we assess the impact of watermarking on face recognition performance. \textbf{Table\,\ref{tab: pretrained_results}} illustrates the recognition performance on two face image datasets: CASIA-WebFace real face images and DCFace synthetic face images. As we can see, watermarking face images results in a marginal decrease in face recognition performance, as measured by True Accept Rate (TAR) at a False Accept Rate (FAR) of 0.01\%. For instance, the average performance drop from `original-original' to `watermarked-original' across the three matchers on the CASIA-WebFace dataset was 0.51\%. This decline in performance was more pronounced when both the probe and reference face images were watermarked, resulting in a TAR drop of 0.81\%. Furthermore, the impact of watermarking was more substantial for the synthetic DCFace dataset, leading to an average decrease of 1.15\% when the probe image was watermarked and 3.35\% when both the probe and reference images were watermarked. The heightened drop in performance on synthetic images due to the watermark raises concerns regarding the use case of applying a watermark to generated face images to maintain provenance, highlighting the need for further methods to embed the watermark without impeding face recognition matching performance.


\begin{figure}[t]
\includegraphics[width=\linewidth,height=!]{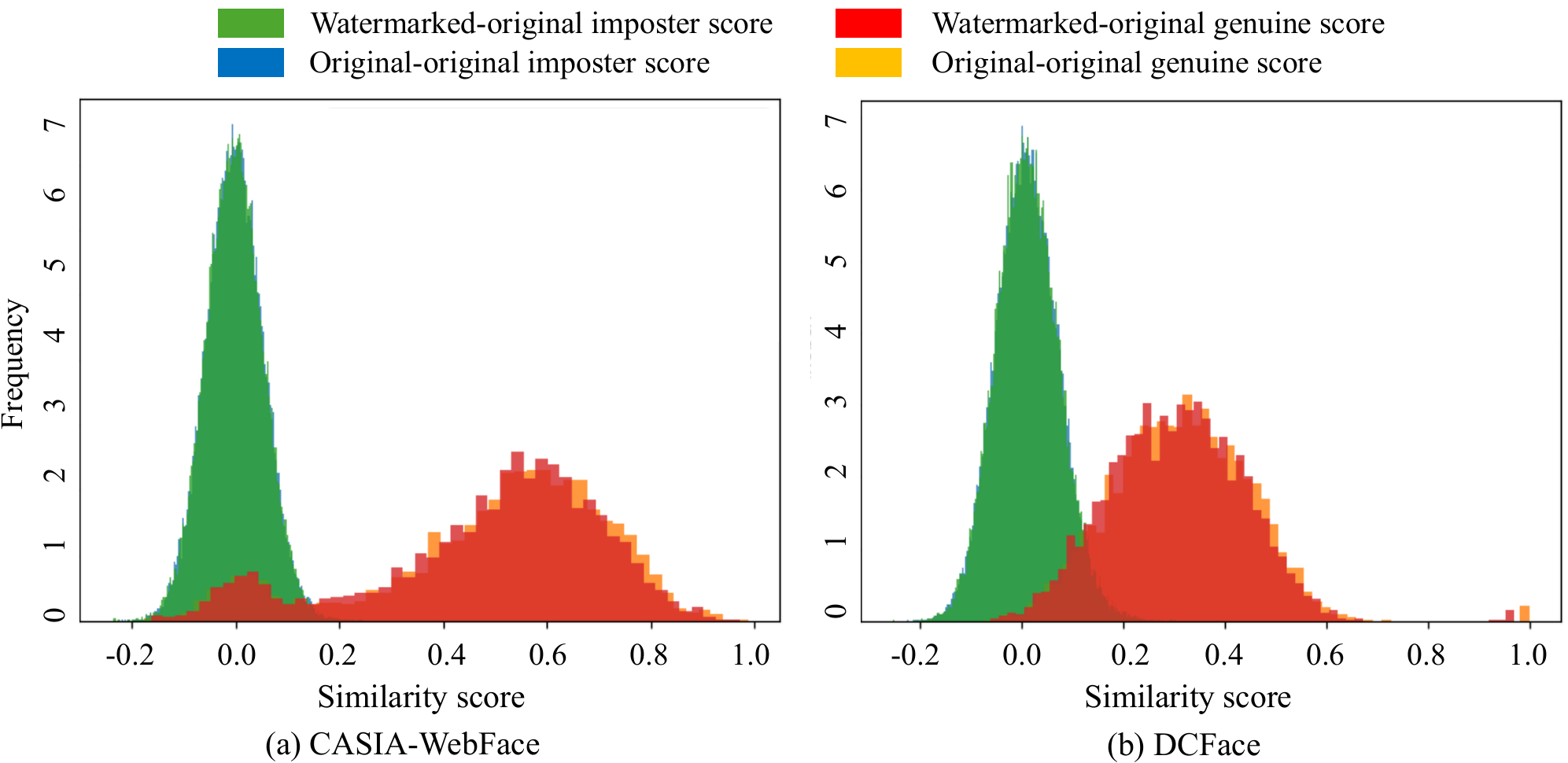}
\vspace{-5mm}
\caption{{ResNet101 AdaFace model pre-trained on MS1Mv2 evaluated on (a) CASIA-WebFace  and (b) DCFace.
}}
\vspace*{-3mm}
\label{fig: hists}
\end{figure} 

\begin{figure}[t]
\includegraphics[width=\linewidth,height=!]{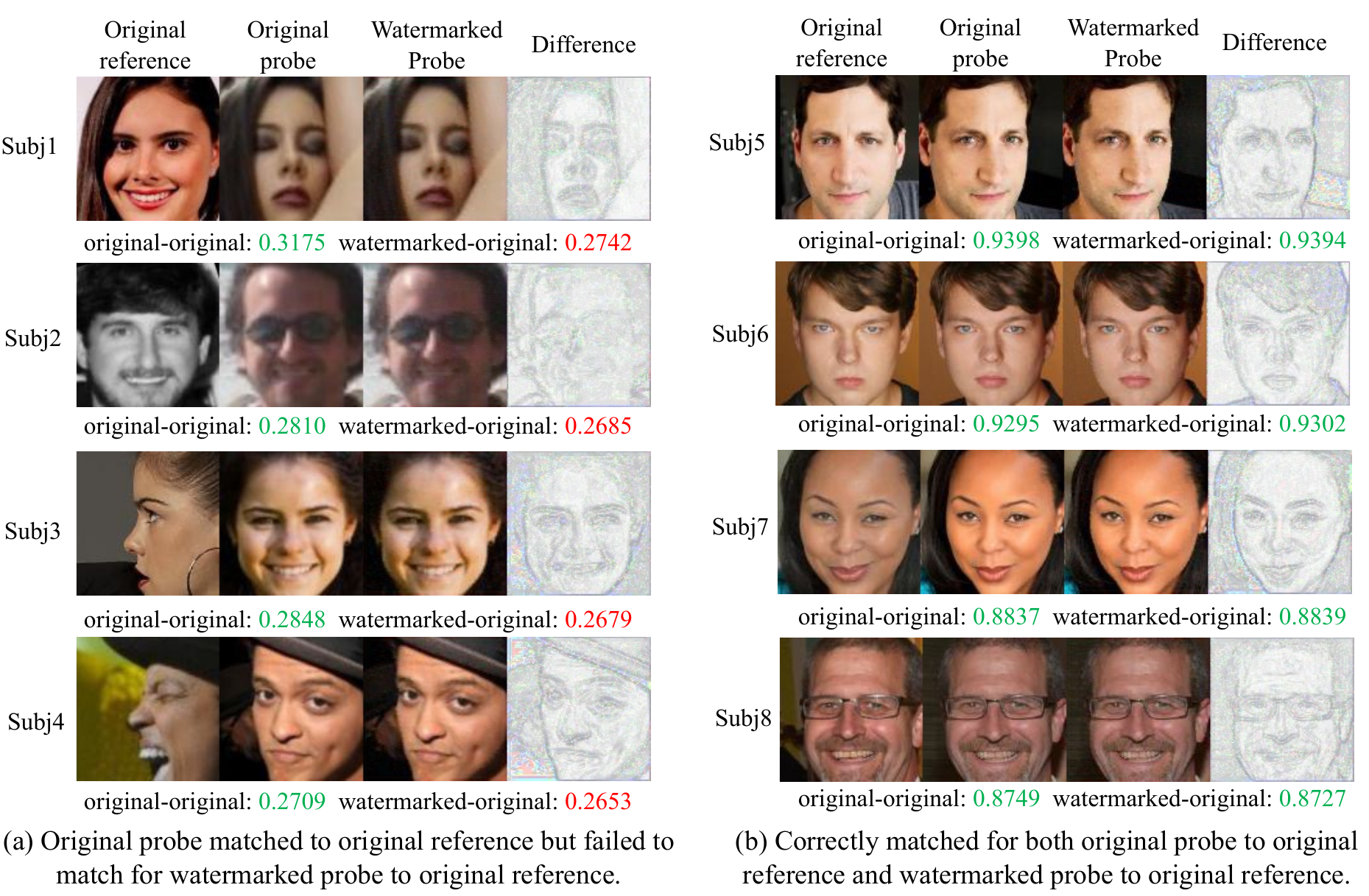}
\vspace{-5mm}
\caption{{Examples of (a) success and (b) failure cases using pre-trained ResNet101 AdaFace on CASIA-WebFace (similarity score is shown for each pair).}}
 \vspace*{-3mm}
\label{fig: failures_and_successes}
\end{figure} 

To better understand the effect of watermarking on face recognition performance, \textbf{Fig.\,\ref{fig: hists}} presents the genuine and imposter similarity score distributions for both CASIA-WebFace and DCFace datasets using the pre-trained ResNet101 AdaFace model before and after applying the watermark to each probe image. As we can see, the imposter score distribution is statistically unchanged whereas the mean of the genuine score distribution is slightly shifted left (from 0.502 to 0.491 for CASIA-WebFace and from 0.313 to 0.303 for DCFace) with the inclusion of the watermarked images. As a result, some challenging comparisons near the match threshold for the original images may be pushed to below the genuine match threshold when watermarked, slightly degrading the performance of the face recognition model. This observation is further supported by examples of the failure cases shown in \textbf{Fig.\,\ref{fig: failures_and_successes}}. These examples showcase challenging genuine comparisons that were matched slightly above the threshold for the original images but resulted in false rejections when the probe image was watermarked.
And these rejections appear to align with matching against challenging poses and expressions.
To validate the statistical significance of the effect of watermarking on the genuine distributions, we conducted a two-sided T-test between the genuine distributions for both datasets before and after applying the watermark. Both results were statistically significant, yielding a p-value of $1.45 \times 10^{-4}$ and $7.15 \times 10^{-8}$ for CASIA-WebFace and DCFace, respectively.


\begin{table}[htb]
\caption{{Effect of watermarking on face recognition training. Trained model performance, in terms of TAR (\%) at FAR = 0.01\%, evaluated on CASIA-WebFace and CASIA-WebFace of watermarked images.
}}
\vspace*{-0mm}
\centering
\resizebox{0.9\linewidth}{!}{%
\begin{tabular}{c|c|c}
\toprule[1pt]
\midrule
\diagbox[]{Train}{Evaluation}
 & \begin{tabular}[c]{@{}c@{}}CASIA-WebFace\\ (original)\end{tabular} & \begin{tabular}[c]{@{}c@{}}CASIA-WebFace\\ (watermarked)\end{tabular} \\
\midrule
\begin{tabular}[c]{@{}c@{}}DCFace\end{tabular} & 54.26 & 49.30 \\
\begin{tabular}[c]{@{}c@{}}DCFace (watermarked)\end{tabular}  & 53.64 & 51.67 \\
\midrule
\begin{tabular}[c]{@{}c@{}}MS1Mv2\end{tabular} & 84.34 & 83.35 \\
\begin{tabular}[c]{@{}c@{}}MS1Mv2 (watermarked)\end{tabular} & 83.69 & 83.12 \\
\midrule
\bottomrule[1pt]
\end{tabular}
}
\label{tab: training_results}
\vspace*{-3.5mm}
\end{table}

\paragraph{Effect of watermarking on the utility of face images for training face recognition models.}

In an ideal scenario, watermarking face images, whether real or synthetic, should not impact downstream face recognition tasks. To evaluate the utility of watermarked face images, we trained multiple face recognition models on both non-watermarked and watermarked versions of the MS1Mv2 and DCFace datasets. We then compared the accuracy of the trained models on the CASIA-WebFace dataset. The resulting face recognition performance is presented in \textbf{Table\,\ref{tab: training_results}}. As we can see, there exists a slight decrease in recognition performance when training on watermarked face images compared to training on the corresponding original face images. For instance, training on MS1Mv2 with and without the watermark led to a drop in recognition accuracy on CASIA-WebFace from a TAR of 84.34\% to 83.69\%. Similarly, a similar trend was observed when training on the DCFace image dataset and its watermarked version, reducing the TAR on CASIA-WebFace from 54.26\% to 53.64\%. 


In the previous experiment (Table\,\ref{tab: pretrained_results}), we noted a decrease in the performance of pre-trained recognition models evaluated on the watermarked version of CASIA-Webface vs. non-watermarked CASIA-Webface. Interestingly, from Table\,\ref{tab: training_results}, we can see that besides the drop in absolute performance compared to training on non-watermarked images (row 1 vs. 2 and row 3 vs. 4), even if the recognition model is trained on watermarked images (as is the case for rows 2 and 4), there is still a gap at test time evaluating on the watermarked version of CASIA-WebFace compared to the non-watermarked CASIA-WebFace (comparing column 2 to column 3); Albeit, this gap is smaller compared to the models only trained on non-watermarked images (rows 1 and 3). Therefore, not only does training on watermarked images decrease the absolute performance of the face recognition model, it also does not provide total robustness to watermarks during test time.

\section{Conclusion}
In this study, we conducted a thorough examination of the effects of digital watermarking on face recognition, emphasizing the complexities and potential benefits associated with the ethical utilization of generated face images. Our devised framework, amalgamating face generation, watermarking, and recognition, facilitated a methodical exploration of their interconnected dynamics. Through rigorous experimentation, we showcased the efficacy of the watermarking scheme in ensuring robust face image attribution, while also shedding light on the inherent trade-offs between watermarking and recognition accuracy. Our results highlight the imperative for future research endeavors aimed at devising techniques capable of optimizing both watermarking and face recognition performance concurrently. Furthermore, the adversarial robustness of watermarking remains inadequately explored and warrants increased research attention in the future. While our study concentrated on assessing the robustness of watermarking against image transformations, its susceptibility to adversarial attacks remains an open concern. Adversarial attacks could potentially render the watermark removable, underscoring the necessity for further investigation in this area.


\section{Broader Impacts}
Our investigation into the impact of digital watermarking on face recognition  holds broader implications for the responsible advancement of AI systems, particularly in the domain of face recognition and biometrics, as well as in the broader realm of generated AI. With the escalating sophistication of generative models in producing lifelike face images, there arises a pressing need to establish measures ensuring the integrity, authenticity, and accountability of these generated images. Exploration in this work could shed light on the potential of digital watermarking as a pivotal tool for image attribution and ownership verification, pivotal in combatting the misuse of generated images for nefarious purposes like deepfake creation and identity fraud. Nevertheless, our findings also underline the imperative of carefully balancing the trade-offs between watermarking and recognition performance. While watermarking offers advantages in terms of image accountability, its presence may impede the accuracy of face recognition systems under certain circumstances. This highlights the necessity of developing techniques that effectively reconcile these competing objectives, ensuring that the benefits of watermarking can be realized without compromising the performance of downstream applications. 
 We aspire that our findings will catalyze further research and stimulate discussions on how the responsible and ethical design of generative AI contribute to broad AI applications.

{\small
\bibliographystyle{unsrt} 
\bibliography{ref/diffusion,ref/ref}
}

\end{document}